\documentclass[runningheads]{llncs}
\usepackage[T1]{fontenc}
\usepackage{graphicx}
\usepackage{amsmath}
\usepackage{amssymb}
\usepackage{booktabs}
\usepackage{siunitx}
\usepackage{etoolbox}
\usepackage{comment}
\usepackage[hidelinks]{hyperref}
\usepackage[nameinlink,capitalize]{cleveref}

\newcommand{\etal}{\textit{et al}.}
\begin{document}
\title{Was that so hard?
Estimating human classification difficulty}

\author{Morten Rieger Hannemose\thanks{These authors contributed equally}\inst{1}%\orcidID{0000-1111-2222-3333} 
\and Josefine Vilsbøll Sundgaard\textsuperscript{\thefootnote}\inst{1}%\orcidID{1111-2222-3333-4444}
 \and Niels Kvorning Ternov\inst{2}%
 \and Rasmus R. Paulsen\inst{1}%\orcidID{2222--3333-4444-5555}
 \and Anders Nymark Christensen\inst{1}%
 }
\authorrunning{Hannemose et al.}
\institute{Department of Applied Mathematics and Computer Science, Technical~University~of~Denmark, Kgs. Lyngby, Denmark \and
Department of Plastic Surgery, Copenhagen University, Herlev~and~Gentofte~Hospital, Copenhagen, Denmark}
\maketitle              % typeset the header of the contribution
\begin{abstract}
When doctors are trained to diagnose a specific disease, they learn faster when presented with cases in order of increasing difficulty. This creates the need for automatically estimating how difficult it is for doctors to classify a given case. In this paper, we introduce methods for estimating how hard it is for a doctor to diagnose a case represented by a medical image, both when ground truth difficulties are available for training, and when they are not. Our methods are based on embeddings obtained with deep metric learning. Additionally, we introduce a practical method for obtaining ground truth human difficulty for each image case in a dataset using self-assessed certainty. We apply our methods to two different medical datasets, achieving high Kendall rank correlation coefficients, showing that we outperform existing methods by a large margin on our problem and data.

\end{abstract}

\begin{figure}[t]
  \centering
    \newcommand{\difffig}[1]{%
    \begin{minipage}{.193\textwidth}%
    \includegraphics[width=\textwidth]{figures/Skin#1.png}\\[1.2pt]%
    \includegraphics[width=\textwidth]{figures/Ear#1.png}%
    \end{minipage}
    }
    \difffig{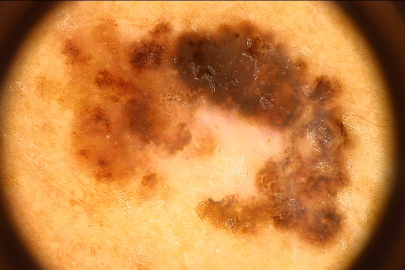}%
    \difffig{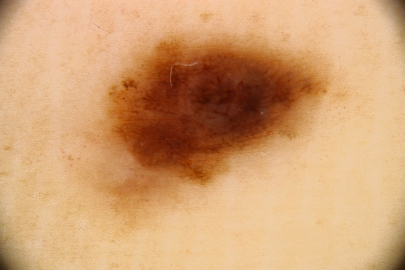}%
    \difffig{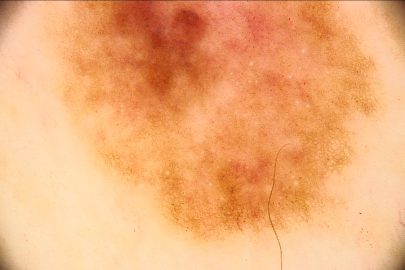}%
    \difffig{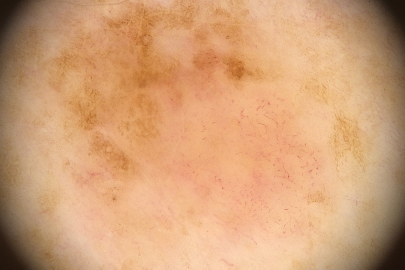}%
    \difffig{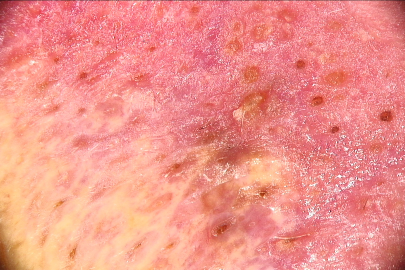}%
   \caption{Image examples from the skin lesion (top row) and eardrum (bottom row) datasets. The difficulty increases from left to right from 0 to 1 in steps of 0.25 for each image. For the skin lesion dataset, only images from the melanoma class are shown, while the eardrum images are from all three diagnostic classes, see \cref{sec:data}.%
   }
   \label{fig:example_images}
\end{figure}

\section{Introduction}
\label{sec:intro}
When doctors are diagnosing patients, not all cases have the same difficulty. A case can be very easy if there are clear diagnostic signs. However, if the typical signs are missing or give conflicting information, a doctor will be more likely to assign an incorrect diagnosis. When doctors are trained to diagnose certain diseases, they learn faster when starting with easy cases and then gradually progressing to harder cases~\cite{roads2018easy}. Knowing how hard each case is to classify is thus useful in an educational context. This concept is well-known in pedagogy~\cite{elio1984effects} and applies to many other areas such as language training, mathematics, etc. 

In this paper, we present a novel approach for estimating human difficulty in image classification using deep metric learning. In deep metric learning, high-dimensional data (in our case, images) are mapped to a lower-dimensional embedding that captures similarities between the training examples: Similar images cluster together, and dissimilar images are pushed apart. In our paper, we define metrics in the embedding space that capture human classification difficulty. We evaluate our methods on two different medical datasets, one containing images of skin lesions and the other of eardrums, see \cref{fig:example_images}.

The term \emph{difficulty} is used in various ways in image analysis. Difficulty can be defined as how hard it is for machine learning to reach high accuracy on a given dataset~\cite{scheidegger2021efficient}, how challenging it is to automatically segment an image~\cite{liu2011estimating}, visual complexity and clutter in the image~\cite{nagle2020predicting}, the time needed for a human to segment an image~\cite{vijayanarasimhan2009s}, or the human response time for a visual search task~\cite{tudor2016hard}. The latter definition was employed by Ionescu \etal~\cite{tudor2016hard}, who proposed a method based on a pretrained neural network for feature extraction, followed by support vector regression to estimate the difficulty score. They presented a dataset with difficulty scores on the PASCAL VOC2012 dataset evaluated by 736 raters. Ma \etal~\cite{ma2019estimating} presented an approach on the same dataset using an end-to-end multi-loss network trained to optimize Kendall's $\tau$ coefficient to predict the difficulty scores. Both approaches achieved high Kendall's $\tau$ coefficients of 0.472 and 0.476, respectively. In contrast to our approach, neither of these use any knowledge about the ground truth class of the image but instead estimated the difficulty directly from image features. By using both the ground truth class and an embedding space our approach becomes more interpretable~\cite{sanakoyeu2019divide}.

In some clinical problems, a specific difficulty scale is already defined. Yoo \etal~\cite{yoo2021deep} predict the difficulty of extracting a mandibular molar from a panoramic radiographic image using a pretrained convolutional neural network. In their study, the Pederson difficulty score is used, which is a pre-defined difficulty scale for extracting mandibular molars. However, this score is purely related to the difficulty of performing the required procedure and not the difficulty of diagnosis. 
André \etal~\cite{andre2010image} propose a method to estimate interpretation difficulty in endomicroscopy videos. Their approach is based on the content-based video retrieval method known as bag-of-visual-words, and the difficulty is given by the percentage of false diagnoses among annotators compared to a ground truth diagnosis from biopsies. We define human difficulty, similarly to André \etal~\cite{andre2010image}, as the fraction of incorrect classifications from people familiar with the classification task. For one of our datasets, we use a self-evaluated certainty of all raters to obtain a less noisy estimated difficulty ground truth with few annotators. 

We compare our work to methods in active learning and curriculum learning. Active learning accelerates labeling efficiency by selecting the most useful samples from an unlabeled dataset for labeling, thus reducing the labeling cost~\cite{ren2020survey}. The intuition behind the most commonly used approach, the uncertainty-based approach, is that with a lower certainty on a specific example, a higher amount of informativeness will be added to the classifier when utilizing the example for training~\cite{wu2020multi}. Curriculum learning is inspired by the learning process of humans, where examples are presented with increasing order of difficulty. This concept is transferred to neural networks to increase training speed and performance by introducing easy examples at the beginning of training, and to gradually increase the difficulty of the training examples~\cite{bengio2009curriculum}.

In this paper, we present a new procedure for obtaining ground truth human difficulty from several annotators by including a self-evaluated certainty. We also propose a new method for estimation of human difficulty based on embeddings of images learned using deep metric learning, which outperforms existing methods by a large margin. We propose methods that both utilize ground truth difficulties and methods that do not. Finally, we are the first to utilize the ground truth class label for human difficulty estimation, which increases the performance of our methods even further.

\section{Estimating image difficulty}
Our difficulty estimation models are all based on the embedding space learned using a deep neural network, trained using metric learning. By training a model this way, instead of as a classification network, we learn the similarities in the training dataset. The output from the network is an embedding vector, mapping each individual image to the embedding space. The idea is that easy cases will be placed far from decision boundaries in the embedding space, while difficult cases will be further away from the class cluster center, and possibly closer to other cluster classes.
We separate our proposed methods into two categories depending on whether or not they utilize ground truth difficulties during training. An overview of these methods is in \cref{tab:performance}.

\textbf{Methods without ground truth difficulties} are all based on embeddings of samples, extracted using a trained neural network. 
As our neural networks are trained using cosine similarity, our methods for estimating difficulties are thus also based on cosine similarities. As difficulties should be high for points far from their cluster, we refer to inverse similarity which is one minus the similarity. The methods still apply to neural networks trained using Euclidean distances, and in that case, one would use the Euclidean distance in the embedding space instead.

\textit{Inverse similarity} is a naïve approach to estimating the difficulty, found by taking the similarity between the sample and the cluster center of its ground truth class. This is intuitive, as samples less similar to the cluster center are typically more similar to other class clusters, and thus harder to classify. To find the difficulty, and not the easiness, we report the inverse of the similarity.

\textit{Inverse softmax of similarity} is an improvement of inverse similarity. Samples can have low similarity to their cluster center without being close to other classes. To handle this, we compute the similarity between the sample and all cluster centers and normalize these with softmax. The difficulty is the inverse of the softmax output corresponding to the ground truth class. 
This method is related to decision margin sampling in active learning~\cite{tuia2011survey}, except we can go on both sides of the decision boundary since the ground truth label is known.

\newcommand{\rocauc}{Sample classification power}
\textit{\rocauc} is an alternative way of obtaining an estimate of image difficulty. Here, we evaluate how many of the neighboring points in the embedding space belong to the ground truth class of a certain sample. To do that for a single sample $s$ from class $c$, we imagine classifying the closest $k$ samples as $c$, and classifying the rest as not $c$. By varying $k$ from one to the number of samples, we can draw a receiver operating characteristic (ROC) curve. We then use the area under the curve (AUC) of this ROC curve as our estimate of the difficulty of $s$. To handle class imbalance, we use the weighted ROC curve, with the weights being the inverses of the class frequencies.

\textit{Normalization} is carried out on the estimated difficulties, by introducing the assumption that each class has the same average difficulty. To enforce this assumption, we propose normalizing the difficulty on a per-class basis by dividing it by the average estimated difficulty of that class. We refer to this as ``norm''.

\textbf{Methods with ground truth difficulties} are methods, where the ground truth difficulties of a training set are employed. We set up a regression problem to predict the difficulty scores directly from the pre-trained image embeddings. We employ the tree-based ensemble model extra trees~\cite{geurts2006extremely} for the regression problem. In addition to only predicting from the embeddings, we also fit a model using the ground truth label as additional input. The ground truth label will allow the model to learn that samples placed close to incorrect class clusters should have a higher difficulty, than samples within their correct class cluster.

\section{Datasets}\label{sec:data}
To validate our method, we have performed experiments on two medical image datasets, examples of which are shown in \cref{fig:example_images}. We have obtained estimates of the human difficulty for a number of images from both datasets, which we use as our test-sets for evaluating our proposed approaches.

\textbf{The skin lesion dataset} consists of dermoscopic images of skin lesions divided into eight diagnoses, which include benign (nevus [NV], keratoses [BKL], vascular lesions [VASC], dermatofibromas [DF]), pre-malignant (actinic keratoses), and malignant (melanoma [MEL], squamous cell carcinoma [SCC], basal cell carcinoma [BCC]).
The diagnoses were determined by histopathology or as the consensus between two to three domain experts.
We have a dataset of 52~292 images from the 2019 ISIC Challenge training set~\cite{tschandl2018ham10000,codella2018skin,combalia2019bcn20000} \footnote{License: CC-BY-NC}
 and our own dataset (Permission to access and handle the patients’ data was granted by the Danish Patient Safety Authority (Jr.\# 3-3013-2553/1) and the Data Protection Agency of Southern Denmark (Jr.\# 18/53664)).

Skin lesion difficulties are obtained for 1723 images from our own dataset, based on diagnoses from 81 medical students with an interest in dermatology (Ethical waiver: Jr.\#: H-20066667, data handling agreement case \#: P-2019-556). On average, each student diagnosed 609 randomly sampled images. It was ensured that at least eight students diagnosed each case.

The images were diagnosed into seven different categories, as we expected actinic keratoses would be too difficult for the medical students. We estimate the difficulty of a case as the fraction of students answering incorrectly. 

\textbf{The eardrum dataset} contains 1409 images collected during the standard clinical routine at an Ear-Nose-and-Throat (ENT) clinic. The data was collected under the ethical approval from the Non-Profit Organization MINS Institutional Review Board (ref.\# 190221). The images show the patients' eardrum captured using an endoscope and are diagnosed into three different diagnoses: acute otitis media, otitis media with effusion, and no effusion by an experienced ENT specialist. The dataset is split into a training and test set of 1209 and 204 images.

Eardrum difficulties were estimated by getting the test set of 204 equally class sampled eardrum images analyzed and diagnosed by four additional experienced ENTs. The ENTs diagnosed each case as one of the three diagnoses or ``unknown'', counting as an incorrect diagnosis. Furthermore, each ENT rated their certainty of each diagnosis on the scale: very low, low, medium, moderate, or high, which is converted to a scale from 0 to 1. More details on this dataset are in Sundgaard \etal~\cite{sundgaard2022inter}.
For a case, $\mu_{correct}$ is the fraction of correct ENT answers and $\mu_{certainty}$ is the average self-evaluated certainty. The difficulty of each case is then
\begin{equation}
    1 -  \mu_{correct} \cdot \mu_{certainty}.
\end{equation} 
We evaluate the difficulties with ``leave-one-annotator-out''. This gave an average Kendall's $\tau$ of 0.548 based only on the fraction of correct ENT answers, which increased to 0.570 when including the self-evaluated certainty, showing that this improves the estimated difficulties.

\section{Experiments}
\label{sec:experiments}
The embeddings of the images in our proposed methods are computed using neural networks trained with a metric loss function. All experiments are conducted in PyTorch (v. 1.10) using the PyTorch metric learning library \cite{musgrave2020pytorch}. The neural networks are trained using the multi-similarity loss function \cite{wang2019multi} ($\alpha$ = 2, $\beta$ = 50, base = 1) and a multi similarity miner ($\epsilon = 0.1$) using cosine similarity to optimize the selection of training pairs. Our models are pretrained on the ImageNet database~\cite{deng2009imagenet} and trained using the Adam optimizer~\cite{kingma2014adam}. The fully connected layer before the final softmax of the model is replaced by a fully connected layer without an activation function, which returns the embedding space. The output embeddings are L2 normalized.

The skin lesion network is based on a ResNet-50 model~\cite{he2016deep}, with a 64-dimensional embedding space. The model is trained for 350 epochs with a learning rate of $10^{-5}$. The input images ($256\times 256$) are color normalized using the Minkowski norm ($p=6$). Data augmentation consists of flips, rotations, scaling, and color jitter. We do inference with the same augmentations, and compute each prediction as the average of 64 random augmentations.

The eardrum network is based on the Inception V3 network~\cite{szegedy2016rethinking}, with a 32-dimensional embedding space. The Inception V3 network has been used by several others for similar images~\cite{cha2019automated,sundgaard2021deep}.
The parameters of the first half of the network (until first grid size reduction) were frozen to avoid over-fitting. 
The initial learning rate ($10^{-3}$) is decreased by a factor of 0.1 every 50\textsuperscript{th} epoch. Training is continued until the training loss has not decreased for 20 epochs, resulting in 111 training epochs. Data augmentation consists of horizontal flips, rotations, color jitter, and random erasing. Images are resized to $299\times 299$.

We use Kendall's~$\tau$ \cite{kendall1948rank} to evaluate how well our methods can predict the ground truth difficulties. This is a non-parametric measurement of the correlation between two ranked variables. As it only compares how the images are ranked, it is not important to achieve the exact same difficulty as the ground truth estimate, as long as the ordering of samples is correct.

We use Extra trees~\cite{geurts2006extremely} for supervised difficulty estimation, with five-fold cross-validation. This allows us to obtain predictions for all samples in the test set, and thus compute a single Kendall's $\tau$ for the entire test set.
All our experiments with extra trees use 500 trees, with 10 as the minimum number of samples required to split an internal node.\footnote{\label{note:scikit}Unspecified parameters are the defaults in Scikit-Learn v. 0.24.2~\cite{scikit-learn}.}

\textbf{Comparisons} are made between our methods and methods from both active and curriculum learning using a standard trained classification network, and with the approach proposed by Ionescu \etal~\cite{tudor2016hard}. The classification networks employ the same architecture as our embedding networks, but the dimension of the output is the number of classes in each dataset. The networks are trained with cross-entropy loss weighted by the inverse frequency of each class, but otherwise using the same setup as described for the embedding networks. 

Visual search difficulty proposed by Ionescu \etal~\cite{tudor2016hard} is used for comparison. We replicate their method by passing each image ($299 \times 299$) through VGG-16 \cite{simonyan2014very} once and using the penultimate features to fit a $\nu$-support vector regression.\textsuperscript{\ref{note:scikit}}

We compare to the following approaches from active learning, all based on the softmax output of a classification network: classification uncertainty, which is one minus the maximum value of the softmax~\cite{lewis1994sequential}; entropy of the softmax probabilities~\cite{Dagan95committeebasedsampling}; and classification margin found by computing the difference between the second-highest and highest probabilities of the softmax~\cite{li2013active}. 

We also compare to three approaches from curriculum learning: standard deviation of the images~\cite{sadasivan2021statistical}; transfer scores obtained by running all images through a pretrained Inception V3 network using the penultimate features to train a support vector classifier to obtain the confidence of the model~\cite{hacohen2019power}; and one minus the softmax output of the ground truth class from our classification network~\cite{hacohen2019power}. 

\section{Results}
The Kendall's~$\tau$ for all experiments is reported in \cref{tab:performance}. The table also gives an overview of whether the ground truth label is used for prediction, and whether a training set of ground truth difficulties has been used.
The embeddings for the two datasets are shown in the top of \Cref{fig:tsne_scatter}. For the eardrum data, we see how most easy examples are located within the class clusters, while the difficult examples are the ones located in another class cluster, or at the edge of the clusters. The same tendencies are visible in some classes of the skin lesion embeddings. The bottom of \Cref{fig:tsne_scatter} shows scatter plots of the ground truth difficulties versus predicted difficulties for both datasets. 

\sisetup{detect-all, table-format=-1.3, input-symbols = {*}, table-space-text-post = {*}}
\newcommand{\onlydiff}{&D}
\newcommand{\onlylabel}{L&}
\newcommand{\both}{L&D}
\newcommand{\neither}{\hspace{12pt} & \hspace{12pt}  }
\newcommand{\somespace}{\quad}
\robustify\bfseries

\begin{table*}[t]
  \caption{Kendall's $\tau$ 
  for all methods on both datasets. It is indicated which methods utilize: the ground truth class label for prediction (L) or a training set of ground truth difficulties (D). 
  Bold indicates the significantly best performance, and bold with a star indicates the methods without D performing significantly better than the rest. We used bootstrap based hypothesis testing with 50~000 replicates and $\alpha = 5\%$.}
  \centering
  \begin{tabular}{@{}lccSS}
    \toprule
    Method &
    \multicolumn{2}{c}{Uses} &
      {\ Skin lesion} & {Eardrum} \\
    \midrule
    \somespace Visual search difficulty \cite{tudor2016hard} & \onlydiff & 0.142 & 0.117 \\    
    \textbf{Curriculum learning} \\
    \somespace Std.\ of image \cite{sadasivan2021statistical} & \neither & -0.070 & 0.011 \\
    \somespace Transfer scoring \cite{hacohen2019power} & \onlylabel & 0.115 & 0.213 \\
    \somespace Self-taught scoring \cite{hacohen2019power} & \onlylabel & 0.176 & 0.261\\
    \textbf{Active learning}\\
    \somespace Classification uncertainty \cite{lewis1994sequential} & \neither & 0.094 & 0.217  \\
    \somespace Entropy of probabilities \cite{Dagan95committeebasedsampling} & \neither & 0.118 & 0.216 \\
    \somespace Classification margin \cite{li2013active} & \neither & 0.068 & 0.215 \\
    \midrule
    \textbf{Ours}\\
    \somespace Inverse similarity & \onlylabel & 0.137& -0.140 \\
    \somespace Inverse softmax of similarity & \onlylabel & \bfseries 0.239* & 0.354\\
    \somespace Inverse softmax of similarity norm. & \onlylabel & \bfseries 0.239* & 0.380  \\
    \somespace \rocauc{} & \onlylabel & 0.201 & 0.143 \\
    \somespace \rocauc{} norm.  & \onlylabel & \bfseries 0.247* & \bfseries 0.440*\\
    \somespace Extra trees: embeddings & \onlydiff & 0.322 & 0.465  \\
    \somespace Extra trees: embeddings + label & \both & \bfseries 0.398 & \bfseries 0.517 \\  
    \bottomrule
  \end{tabular}
  \label{tab:performance}
\end{table*}

\section{Discussion and conclusion}
We have shown that neural networks trained using metric learning can be used to estimate diagnostic difficulty. Our methods for difficulty estimation outperform all existing methods in both active and curriculum learning.
Ionescu \etal~\cite{tudor2016hard} report a Kendall's $\tau$ of 0.472, while their method achieves 0.142 and 0.117 on our datasets. Our methods are significantly better, with our best achieving a Kendall's $\tau$ of 0.398 and 0.517. This corresponds to 69.9\% and 75.8\% of pairs being ordered correctly, which is an improvement of 12.8 and 11.1 percentage points from the best performing existing method (self-taught scoring). 

\begin{figure*}[t]
\newcommand{\figemph}{\textbf}
  \centering
   \includegraphics[width=\linewidth]{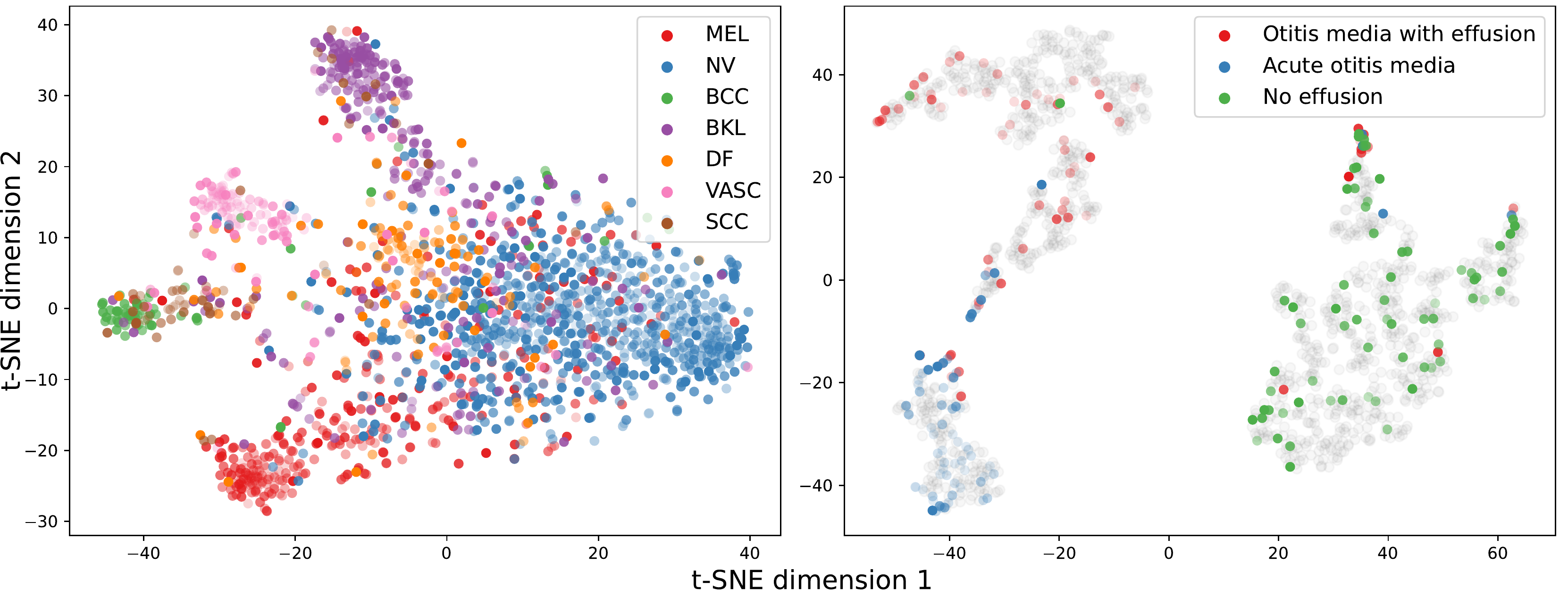}
   \includegraphics[width=\linewidth]{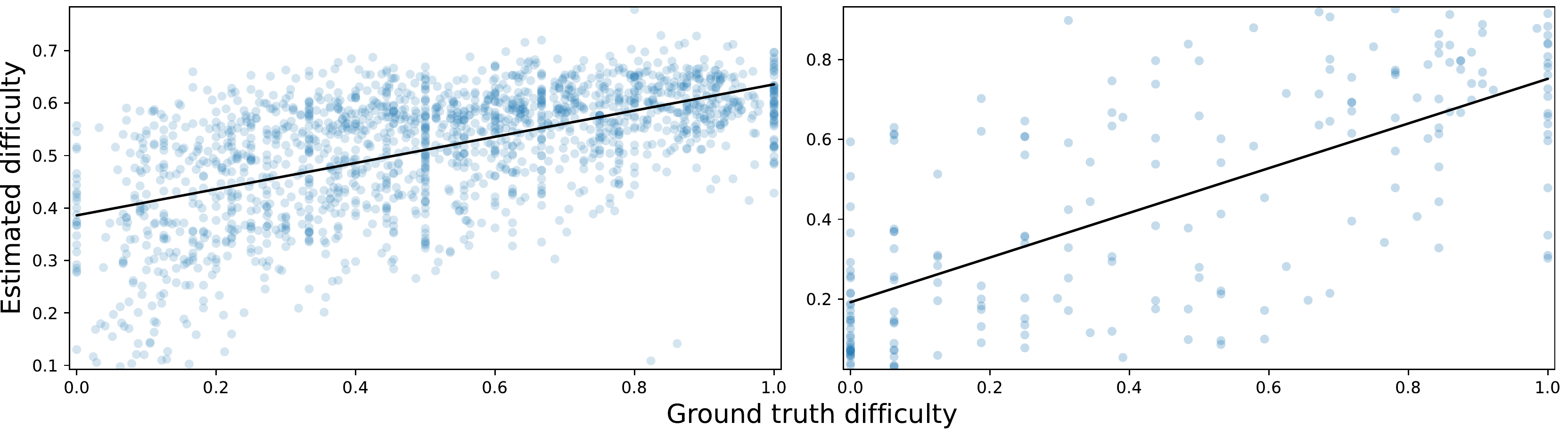}
   \caption{\figemph{Left}: skin lesion dataset. \figemph{Right}: eardrum dataset. \figemph{Top}: visualization of the embeddings in two dimensions with t-SNE \cite{van2008visualizing}.
   The transparency of each point indicates the ground truth difficulty with very transparent being the easiest. Grey points are the training samples for the eardrum data.
   %visualization of embeddings with t-SNE \cite{van2008visualizing}.
   %Transparency of each point indicates ground truth difficulty with very transparent being the easiest. Grey points are training samples for the eardrum data.
   \figemph{Bottom}: scatter plots of ground truth difficulties and difficulties estimated with the \emph{embeddings + label} approach, together with the least squares regression lines.}
   \label{fig:tsne_scatter}
\end{figure*}

\Cref{tab:performance} shows that our contribution of incorporating the ground truth class greatly increases performance. A similar tendency is seen in the higher performance of self-taught scoring compared to classification uncertainty, as the only difference between these two methods is the knowledge about the ground truth class. This intuition is also visible in \cref{fig:tsne_scatter}, especially for the eardrum dataset, where the most difficult examples are often placed in the extremities of the clusters, or placed inside other clusters. This indicates that the embedding has a relation to difficulty, and shows the relevance of including the ground truth class label when estimating difficulty. Our methods have demonstrated great potential in the estimation of human classification difficulty of medical images, which can be used to optimize and improve the training of medical professionals.
\bibliographystyle{splncs04} 
\bibliography{egbib}

\end{document}